\documentclass[11pt]{article}
\usepackage[preprint]{acl}
\usepackage{times}

\usepackage{latexsym}
\usepackage{amsmath,amssymb,booktabs,graphicx,xcolor}
\usepackage{microtype}
\usepackage{float}
\usepackage{flafter}

\newcommand{\method}{\textsc{OPSD4MAS}}
\newcommand{\factfilter}{\textsc{FEHF}}

\newcommand{\DeltaFWTwoVsDAPOSixDatasetMeanPFour}{4.83}

\newcommand{\DeltaFWFourVsGRPOMax}{20.00}

\newcommand{\DeltaFWFourVsGRPOSixDatasetMeanPOne}{8.87}
\newcommand{\DeltaFWFourVsGRPOSixDatasetMeanPFour}{9.54}

\newcommand{\DeltaFWFourVsDAPOSixDatasetMeanPOne}{6.14}
\newcommand{\DeltaFWFourVsDAPOSixDatasetMeanPFour}{8.48}

\newcommand{\DeltaFWFourVsFWTwoSixDatasetMeanPOne}{6.06}
\newcommand{\DeltaFWFourVsFWTwoSixDatasetMeanPFour}{3.65}

\newcommand{\DeltaFWFourVsCoMASSixDatasetMeanPOne}{8.46}
\newcommand{\DeltaFWFourVsCoMASSixDatasetMeanPFour}{11.13}

\newcommand{\AblExtractionPOne}{9.33}
\newcommand{\AblExtractionPFour}{13.52}
\newcommand{\AblAssociationPOne}{0.42}
\newcommand{\AblAssociationPFour}{1.21}
\newcommand{\AblLOOPOne}{21.92}
\newcommand{\AblLOOPFour}{17.56}
\newcommand{\AblFullVsBasePOne}{15.28}
\newcommand{\AblFullVsBasePFour}{16.15}

\newcommand{\IntroInjectionMin}{1.98}
\newcommand{\IntroInjectionMax}{6.13}

\title{Turning Multi-Agent Communication into Token-Level Credit\\with On Policy Self Distillation}

\author{
  Zixu Wang$^{1,*}$ \quad Yu Zhao$^{1,*}$ \quad Xing Zheng$^{3}$ \quad
  Bingbing Xu$^{1}$ \quad Tian Xueyun$^{1}$ \\
  \bfseries JiaTang Luo$^{2}$ \quad Weiqiang Wang$^{3}$ \quad
  Bowen Song$^{3,\dagger}$ \quad Huawei Shen$^{1}$ \quad Xueqi Cheng$^{1,\dagger}$ \\[0.4em]
  {\normalfont\small $^{1}$Institute of Computing Technology, Chinese Academy of Sciences} \\
  {\normalfont\small $^{2}$University of Chinese Academy of Sciences \quad $^{3}$Ant Group} \\[0.3em]
  {\normalfont\small $^{*}$Equal contribution. \quad $^{\dagger}$Corresponding authors.} \\
  {\normalfont\small\texttt{wangzixu22s@ict.ac.cn}} \\
  {\normalfont\small\texttt{bowen.sbw@antgroup.com} \quad \texttt{cxq@ict.ac.cn}}
}

\hypersetup{
  pdftitle={Turning Multi-Agent Communication into Token-Level Credit with On Policy Self Distillation},
  pdfauthor={Zixu Wang, Yu Zhao, Xing Zheng, Bingbing Xu, Tian Xueyun, JiaTang Luo, Weiqiang Wang, Bowen Song, Huawei Shen, Xueqi Cheng}
}
\begin{document}
\raggedbottom
\maketitle
\begin{abstract}
Reinforcement learning improves problem solving in large language
model multi-agent systems.
Existing approaches overlook communication knowledge or compress it into
scalar rewards, leaving coarse credit with little guidance on which decisions
each agent should reinforce or revise.
OPSD4MAS converts future communication into fine-grained token-level credit
through On Policy Self Distillation (OPSD).
An external model extracts facts and assesses their trajectory relations for
coarse-grained filtering, followed by fine-grained leave-one-out filtering of
target-token influence against agent-wise outcome advantages.
The fact-conditioned policy supplies token-level credit, combined with GRPO
through a linearly annealed joint objective.
Our method improves mathematical pass@4 over GRPO by up to
\DeltaFWFourVsGRPOMax{} percentage points.
Ablations support complete filtering, while frozen-history injection
diagnostics support fact utility. Together, these results support combining
agent-level outcome guidance with communication-derived token-level credit.
Code is available at
\url{https://github.com/Camel-Prince/OPSD4MAS}.
\end{abstract}

\begin{figure}[!t]
\input{figures/overview_teaser}
\end{figure}
\section{Introduction}
\label{sec:intro}

Reinforcement learning (RL) improves large language model (LLM) reasoning by
updating policies from feedback on the quality of their generated answers
\citep{shao2024deepseekmath,yu2025dapo}.
Extending RL to LLM-based multi-agent systems (MAS) trains interacting
agents to reason and communicate for joint success
\citep{liu2025magrpo,feng2026drmas}.
Most approaches derive scalar rewards from task outcomes or agent
communication, assign credit to participating agents, and update their
policies to improve multi-turn reasoning and coordination within each task.

Many recent MAS training methods follow \emph{scalar-reward RL}.
Methods such as MAGRPO, M-GRPO, and MARTI turn task outcomes,
intermediate evaluations, or judgments of agent interactions into scalar
rewards for policy updates
\citep{liu2025magrpo,hong2025mgrpo,zhang2026marti}.
Figure~\ref{fig:overview}(a) contrasts two common routes.
\emph{Outcome-derived rewards} assign task-success signals to responses,
as in the agent-wise advantage normalization of
DrMAS \citep{feng2026drmas}.
\emph{Communication-informed rewards} additionally use signals from discussions,
as in MAPoRL and CoMAS, while retaining scalar feedback
\citep{park2025maporl,xue2026comas}.
These scalar-reward approaches either leave communication-derived knowledge
unused in reward construction or compress it into scores, losing a valuable
source of learning signal. For an intermediate response in a MAS, future
communication can explain an error, validate a claim, or suggest a repair.
Scalar rewards also yield coarse credit assignment: a response-level advantage
indicates whether to reinforce a response, but cannot itself distinguish which
tokens a correction supports or challenges, even when \mbox{token gradients differ.}

To address this limitation, we propose OPSD4MAS, which converts future
communication into \emph{token-level credit}
using On Policy Self Distillation (OPSD), in which the current
policy conditioned on selected future information acts as a privileged
teacher for the same policy under its ordinary context. Comparing their
probabilities on student-generated tokens supplies token-level credit,
as shown in Figure~\ref{fig:overview}(a), incorporating communication-derived knowledge
into agent parameters without future information \mbox{at inference time.}

The key to OPSD4MAS is constructing useful privileged context.
Unfiltered communication mixes useful corrections with errors and irrelevant
remarks. We therefore introduce
\emph{Fact Extraction and Hierarchical Filtering} (\factfilter{}).
For each target response, an external model extracts facts from its future
communication window and labels their relations to same-query trajectories.
Association filtering ranks candidate facts by their relevance to solving the
current query, assessed from whether following or violating each fact aligns
with the outcome advantages of trajectories for that query. Facts followed
in higher-advantage trajectories or violated in lower-advantage trajectories
receive higher scores. Leave-one-out (LOO) filtering then checks whether each
candidate fact changes the target response's token probabilities in the
direction favored by that response's GRPO outcome advantage.
Retained facts provide privileged context without external teacher logits.
We combine OPSD with reinforcement learning with verifiable rewards (RLVR),
using GRPO and linearly decreasing the OPSD loss weight.

Across six benchmarks in Table~\ref{tab:main}, OPSD4MAS with a future window
of four responses (FW4) improves mean pass@1 by
\DeltaFWFourVsGRPOSixDatasetMeanPOne{} and \DeltaFWFourVsDAPOSixDatasetMeanPOne{} percentage points
over GRPO and DAPO, respectively. Mean pass@4 gains are
\DeltaFWFourVsGRPOSixDatasetMeanPFour{} and \DeltaFWFourVsDAPOSixDatasetMeanPFour{} percentage points.
Figure~\ref{fig:overview}(b) shows three representative datasets.
We also inject selected facts into frozen histories and regenerate the final
Solver answer. With Relaxed filtering, accuracy exceeds no injection by
\IntroInjectionMin{}--\IntroInjectionMax{} percentage points across four checkpoints
in Figure~\ref{fig:injection-strength}.
The training ablations in Table~\ref{tab:ablations} compare raw communication,
extracted facts, and the two filtering stages. Extraction, Association, and
LOO improve mean pass@1 over their preceding ablated settings by
\AblExtractionPOne{}, \AblAssociationPOne{}, and \AblLOOPOne{} percentage points,
respectively, with the largest gain \mbox{from target-specific LOO.}

Overall, our contributions are threefold:
\begin{itemize}
  \item \textbf{We turn communication into token-level credit.}
  We identify knowledge lost or compressed by scalar-reward MAS training and
  use future communication as privileged context for token-level credit.
  \item \textbf{We develop fact-guided joint OPSD--RLVR training.}
  \factfilter{} combines fact extraction with coarse- and fine-grained filtering.
  The fact-guided teacher supplies OPSD credit alongside RLVR, instantiated
  with GRPO, under a linearly annealed joint objective.
  \item \textbf{We show performance gains and downstream fact utility.}
  OPSD4MAS (FW4) improves both metrics across six benchmarks over all three trained baselines,
  with gains up to \DeltaFWFourVsGRPOMax{} percentage points.
  Ablations and frozen-history injection diagnostics support the filtering
  design and the downstream utility of selected facts.
\end{itemize}

\section{Related Work}

\paragraph{Multi-agent systems.}
LLM-based MAS coordinate specialized agents through role-based communication,
debate, and shared task execution
\citep{li2023camel,wu2023autogen,du2023debate}.
Beyond manually specified interactions, AFlow and ADAS search for effective
agent organizations \citep{zhang2025aflow,hu2025adas}.
These approaches address how agents collaborate. A complementary question is
how post-training can improve the policies participating in that collaboration.

\paragraph{Reinforcement learning for multi-agent systems.}
RL post-training improves LLM policies using feedback on generated answers.
GRPO, RLOO, and DAPO offer policy-gradient approaches with different advantage
estimators and optimization designs
\citep{shao2024deepseekmath,ahmadian2024rloo,yu2025dapo}.
Extending reward-based learning to interacting agents introduces the problem
of assigning credit across roles and turns. MAGRPO optimizes cooperative
policies using shared rewards, while DrMAS normalizes outcome rewards with
agent-specific statistics \citep{liu2025magrpo,feng2026drmas}.
To account for collaboration, MAPoRL shapes verifier rewards using changes in
answer correctness and influence across discussions, while CoMAS derives scalar
action rewards from interaction-based judgments
\citep{park2025maporl,xue2026comas}.
However, these methods reduce task feedback to scalar rewards rather than
use communication content directly for task-specific token-level credit.
OPSD4MAS addresses this gap by conditioning a privileged teacher on selected
communication while retaining outcome credit.

\paragraph{On Policy Self Distillation.}
On Policy Distillation (OPD) is an emerging LLM post-training paradigm that
queries a teacher on student-generated trajectories, providing token-level
learning signals on states visited by the student \citep{song2026opdsurvey}.
In On Policy Self Distillation (OPSD), the student model
or a copy acts as teacher with additional information unavailable to the
ordinary-context student. Related methods differ in their teachers and
privileged information. OPCD studies frozen and shared-weight context teachers
\citep{ye2026opcd}. MOPD uses successful and failed same-question rollouts
with self or external teachers \citep{yu2026mopd}. GEAR uses a ground-truth-conditioned
self-model to reweight outcome advantages, not a direct distillation loss
\citep{li2026gear}. TurnOPD uses a frozen external teacher on student-visited
multi-turn histories \citep{zhou2026turnopd}. EDGE-OPD masks updates by privileged-context
evidence, whereas AR-OPD controls full-view residual guidance around a
partial-privilege anchor
\citep{lazaridis2026edgeopd,zhang2026aropd}.
Our focus is selecting reliable, target-specific facts from mixed-quality
future communication, rather than directly adopting complete answers or
alternative rollouts as privileged context.

\paragraph{OPD for MAS.}
Recent work applies OPD to multi-agent learning.
MAS-OPD supplies a frozen external teacher with verifier-produced attributions
of past-turn conflicts \citep{zhong2026masopd}.
$\pi$-Play conditions an EMA self-teacher on an examiner's question-construction
paths \citep{zhang2026piplay}.
OPSD4MAS instead extracts facts from future inter-agent communication during
task solving. Cross-trajectory association and target-specific LOO filtering
select these facts as privileged context for each role's privileged teacher.

\section{Problem Formulation}

A language-model multi-agent system (MAS) coordinates communicating agents
through a predefined workflow, graph, or autonomous decisions.
In the role-specific setting, each role $r\in\mathcal R$ has parameters
$\theta_r$, and $\Theta=\{\theta_r:r\in\mathcal R\}$ collects all trainable
parameters. We write $\pi_\theta^r=\pi_{\theta_r}$ for its policy.
For query $q$, index sampled trajectories by $j$. \mbox{Stochastic execution gives}
\begin{equation}
\tau_j=\operatorname{MAS}_{\Theta}(q),\qquad
\tau_j=(y_{j,1},\ldots,y_{j,T_j}).
\end{equation}
Response $y_{j,i}$ is generated by role $r(i)$, which can act repeatedly.
Token $t$ has context $(q,y_{j,<i},y_{j,i,<t})$ under shared history.
Restricted communication exposes only the permitted prefix.
MAS RL updates all participating agents' parameters, not only the final
answering agent. With outcome reward $R(\tau)$, the objective is
\begin{equation}
\max_{\Theta}\;\mathbb E_{q,\,\tau\sim\operatorname{MAS}_{\Theta}(q)}[R(\tau)].
\end{equation}
DrMAS \citep{feng2026drmas} normalizes outcome rewards by agent within each
query group to obtain
\begin{equation}
A^{r(i)}_{j,i}=\frac{s_{j,i}-\mu_{q,r(i)}}{\sigma_{q,r(i)}+\epsilon}.
\label{eq:grpo-outcome-advantage}
\end{equation}
Here $s_{j,i}$ is the outcome reward assigned to response $y_{j,i}$ in GRPO
training. Mean $\mu$ and standard deviation $\sigma$ pool all same-agent
responses across trajectories for query $q$, counting repeated turns
separately. This agent-wise advantage weights every valid response token
in GRPO's clipped surrogate.
Communication-informed CoMAS instead optimizes interaction-derived action
rewards with REINFORCE++ and a clipped surrogate \citep{xue2026comas}.
Its parameters index agents that can perform different interaction roles.
Appendix~\ref{app:baseline-objectives} details their normalization and KL terms.

\section{Method}
\label{sec:method}

\begin{figure*}[t]
  \centering
  \includegraphics[width=\textwidth]{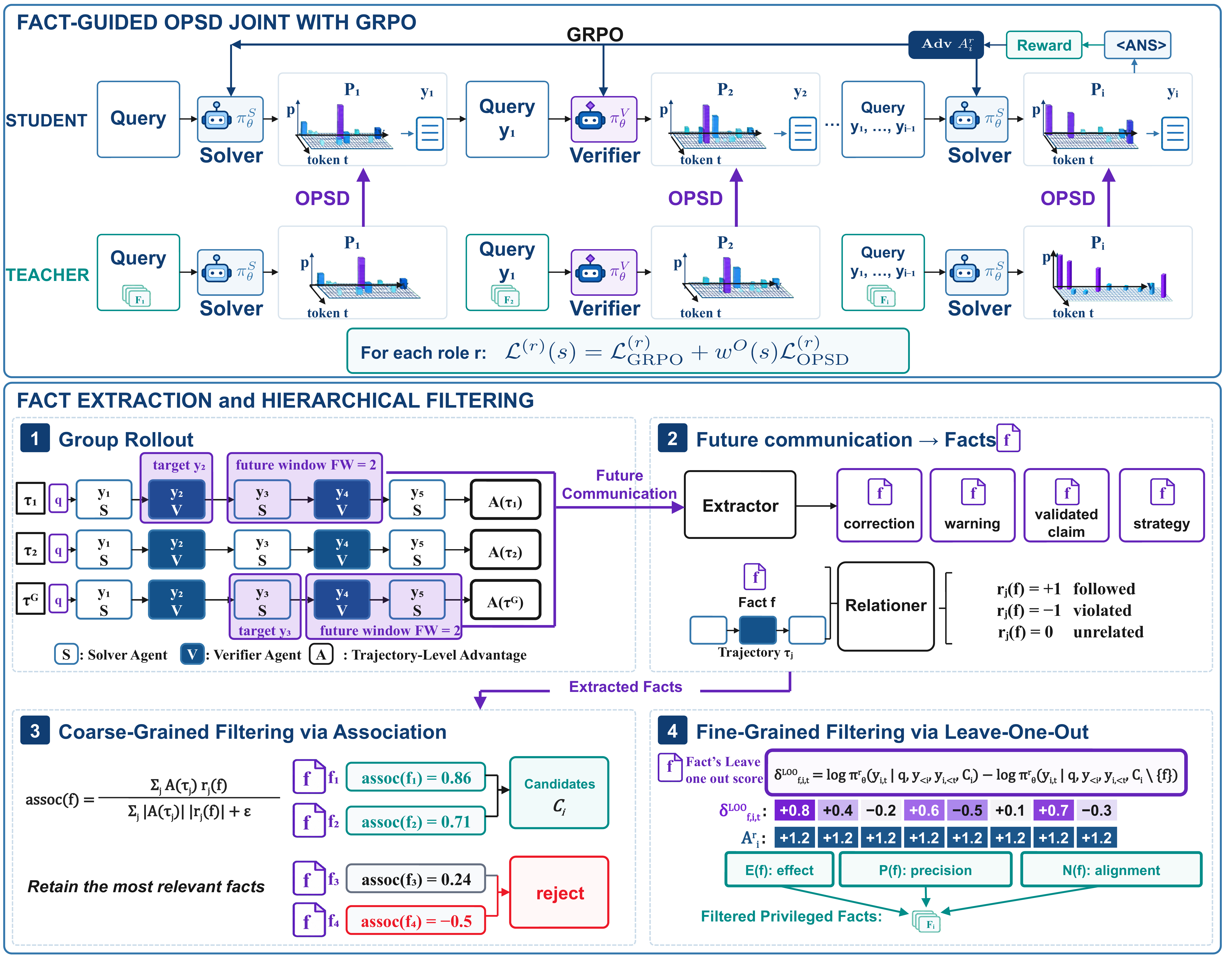}
  \caption{OPSD4MAS combines agent-wise GRPO credit with token-level OPSD
  credit. Future communication is converted into facts, ranked by association
  with query-normalized trajectory advantage $A(\tau_j)$, and filtered by LOO
  against DrMAS agent-wise advantage $A_i^r$, also used by GRPO. The selected set $F_i$ conditions
  each role's privileged teacher. Probability displays and numerical examples are
  illustrative.}
  \label{fig:method}
\end{figure*}

\subsection{Motivation}

Scalar-reward MAS training discards or compresses communication knowledge
useful for collaboration. We use future communication as privileged context
for token-level OPSD credit. Mixing corrections with errors and irrelevant
remarks motivates fact extraction and two filtering stages. Communication
also lacks verified outcome feedback, motivating annealed joint OPSD--GRPO
training. Following Figure~\ref{fig:method}, we describe privileged-context
construction, then joint training.

\subsection{Fact Extraction and Hierarchical Filtering}

\paragraph{Fact extraction.}
Given trajectory $\tau_j$, consider its $i$-th response $y_i$, omitting $j$
below and writing $r=r(i)$. An external model examines communication in
$y_{i+1},\ldots,y_{\min(i+W,T)}$ and extracts atomic corrections, warnings,
validated claims, and strategies (Figure~\ref{fig:method}, panels 1--2).
Facts retain their source and target identities. The terminal response can
supply facts to earlier steps but has no future source itself.
Appendix~\ref{app:fact-prompts} gives the Extractor and Relationer prompts.
Extraction converts communication into individual facts, and two filters then
select the target's privileged information.

\paragraph{Association filtering.}
Association filtering (Figure~\ref{fig:method}, panel 3) uses fact--trajectory
relations for the same query, labeled by the external model:
$r_j(f)\in\{+1,-1,0\}$ for \emph{followed}, \emph{violated}, and
\emph{unrelated}. For fact ranking, we separately compute
$A(\tau_j)=(R(\tau_j)-\mu_q)/(\sigma_q+\epsilon)$,
counting each trajectory once in the query group's reward statistics.
Unlike the response-occurrence normalization in
Equation~\ref{eq:grpo-outcome-advantage}, this does not weight repeated agent turns.
We compute
\begin{equation}
\operatorname{assoc}(f)=
\frac{\sum_j A(\tau_j)r_j(f)}
{\sum_j |A(\tau_j)|\,|r_j(f)|+\epsilon}.
\end{equation}
Following a fact at positive advantage or violating it at negative advantage
contributes positively, whereas the reverse contributes negatively. Unrelated labels
contribute neither term. The score ranks facts by outcome association on the
query rather than weighting the actor loss. Retaining a limited number forms $C_i$,
without a positive-score threshold. Appendix~\ref{app:algorithm} details
normalization, exclusions, and tie-breaking.

\paragraph{Target-specific LOO filtering.}
From the Association-ranked candidates, LOO retains facts with sufficient
influence on the target's tokens, predominantly aligned with its outcome
advantage (Figure~\ref{fig:method}, panel 4). It compares the full candidate
set with one fact removed, using cached pre-update policy scores. Write
\begin{equation*}
g_{i,t}(S)=\log\pi_\theta^r(y_{i,t}\mid q,y_{<i},y_{i,<t},S).
\end{equation*}
We define the removal effect and its outcome-directed value as
\begin{equation}
\begin{aligned}
\delta^{\mathrm{LOO}}_{f,i,t}
  &=g_{i,t}(C_i)-g_{i,t}(C_i\setminus\{f\}),\\
z_{f,i,t}&=A_i^r\delta^{\mathrm{LOO}}_{f,i,t}.
\end{aligned}
\end{equation}
Here $A_i^r$ is the target's DrMAS agent-wise advantage used by GRPO
(Equation~\ref{eq:grpo-outcome-advantage}, $j$ omitted), not Association's
$A(\tau_j)$. Let
$m_+=\mathbb{E}_t[\max(z_{f,i,t},0)]$ and
$m_-=\mathbb{E}_t[\max(-z_{f,i,t},0)]$, where expectations include only active
response tokens. We measure effect, directional precision, \mbox{and net alignment:}
\begin{align}
E(f)&=\sqrt{\mathbb{E}_t[(\delta^{\mathrm{LOO}}_{f,i,t})^2]},\\
P(f)&=\frac{m_+}{m_++m_-+\epsilon},\\
N(f)&=m_+-\lambda_c m_-.
\end{align}
$E$ measures influence on token log probabilities, while $m_+$ and $m_-$ measure
its advantage-aligned and opposing mass. $P$ is the positive influence share,
not a token fraction, and $N$ penalizes opposing mass. Useful facts should have
sufficient influence, a high positive share, and a large net effect.
For this target, a fact passes when $E\ge E_{\min}$, $P\ge P_{\min}$, and
$N>N_{\min}$. Top-$K$ ranking gives the final set $F_i\subseteq C_i$.
We use $\lambda_c=1$ by default and rescore $F_i$ for OPSD.
Context-matching and budget details appear in Appendix~\ref{app:serialized-contrast}.

\begin{table*}[t]
\centering
\small
\setlength{\tabcolsep}{1.7pt}
\begin{tabular*}{\textwidth}{@{\extracolsep{\fill}}l*{12}{r}@{}}
\toprule
Method & \multicolumn{2}{c}{AMC23} & \multicolumn{2}{c}{AIME24} & \multicolumn{2}{c}{AIME25} & \multicolumn{2}{c}{MATH-500} & \multicolumn{2}{c}{Olympiad} & \multicolumn{2}{c}{Minerva} \\
\cmidrule(lr){2-3}\cmidrule(lr){4-5}\cmidrule(lr){6-7}\cmidrule(lr){8-9}\cmidrule(lr){10-11}\cmidrule(lr){12-13}
 & $p_1$ & $p_4$ & $p_1$ & $p_4$ & $p_1$ & $p_4$ & $p_1$ & $p_4$ & $p_1$ & $p_4$ & $p_1$ & $p_4$ \\
\midrule
Base Model & 60.24 & 74.70 & 25.83 & 43.33 & 19.17 & 23.33 & 82.40 & 90.40 & 41.04 & 53.04 & 24.72 & 31.99 \\
\midrule
\multicolumn{13}{@{}l}{\textit{Frozen-weight MAS: no LLM parameter updates}} \\
Solver--Verifier MAS & 56.63 & 71.08 & 22.50 & 36.67 & 16.67 & 26.67 & 83.50 & 91.80 & 46.81 & 59.26 & 33.27 & 41.18 \\
AFlow & 61.44 & 77.11 & 27.50 & 46.67 & 22.50 & 26.67 & 79.15 & 88.60 & 45.44 & 58.07 & 30.61 & 39.34 \\
GPTSwarm & 61.44 & 74.70 & 20.83 & 36.67 & 20.83 & 30.00 & 80.75 & 89.60 & 44.33 & 54.22 & 24.72 & 34.19 \\
\midrule
\multicolumn{13}{@{}l}{\textit{Scalar-reward RL: sparse credit assignment}} \\
DrMAS-GRPO & 68.07 & 83.13 & 48.33 & 63.33 & 32.50 & 53.33 & \underline{\textit{90.95}} & 94.60 & 56.11 & 68.30 & 36.76 & 42.65 \\
DrMAS-DAPO & \underline{\textit{78.01}} & 85.54 & 46.67 & 66.67 & 38.33 & 53.33 & 90.35 & 94.40 & \underline{\textit{58.93}} & 68.00 & 36.86 & 43.75 \\
CoMAS & 71.29 & 81.93 & 43.33 & 60.00 & 36.67 & 50.00 & 90.90 & 94.20 & 56.59 & \underline{\textit{68.44}} & 36.40 & 41.18 \\
\midrule
\multicolumn{13}{@{}l}{\textit{Ours: token-level credit assignment}} \\
\textbf{OPSD4MAS (FW2)} & 74.10 & \underline{\textit{87.95}} & \underline{\textit{50.83}} & \underline{\textit{70.00}} & \underline{\textit{41.67}} & \underline{\textit{73.33}} & 89.00 & \underline{\textit{95.20}} & 55.59 & 67.11 & \underline{\textit{38.42}} & \underline{\textit{47.06}} \\
\textbf{OPSD4MAS (FW4)} & \underline{\textbf{82.23}} & \underline{\textbf{91.57}} & \underline{\textbf{62.50}} & \underline{\textbf{83.33}} & \underline{\textbf{50.00}} & \underline{\textbf{73.33}} & \underline{\textbf{91.50}} & \underline{\textbf{96.60}} & \underline{\textbf{60.30}} & \underline{\textbf{69.93}} & \underline{\textbf{39.43}} & \underline{\textbf{47.79}} \\
\bottomrule
\end{tabular*}
\caption{Test performance (\%): $p_1$/$p_4$ denote pass@1/pass@4. Bold underlined and italic underlined values rank first and second in each column. All nine methods are evaluated on six datasets. Olympiad denotes English text-only OlympiadBench mathematics. Evaluation protocols and scoring qualifications appear in Section~\ref{sec:setup} and Appendix~\ref{app:planned-comparisons}.}
\label{tab:main}
\end{table*}

\subsection{Fact-Conditioned OPSD and Annealed Joint Training}

\paragraph{Fact-conditioned token-level credit.}
The privileged teacher and ordinary-context student are two views of the
same role policy. Using the selected set $F_i$, we compute
\begin{equation}
\begin{aligned}
\delta_{i,t}={}&\log\pi_\theta^r(y_{i,t}\mid q,y_{<i},y_{i,<t},F_i)\\
&-\log\pi_\theta^r(y_{i,t}\mid q,y_{<i},y_{i,<t}).
\end{aligned}
\end{equation}
Scores are cached before actor updates. The token-dependent coefficient is
$A^{\textsc{opsd}}_{i,t}=\operatorname{sg}(\lambda_O\delta_{i,t})$, where
$\operatorname{sg}$ stops gradients and $\lambda_O=1$ by default.
It is zero when no facts are selected. Unlike either outcome advantage,
this coefficient comes from token log-probability differences.

\paragraph{Role-wise OPSD objective.}
Within one role-homogeneous loss-reduction batch, let $I_r$ contain role $r$'s responses, $m_{i,t}$ be the active-token
mask, and $Z_r=\sum_{i\in I_r}\sum_t m_{i,t}$. For $Z_r>0$,
\begin{equation}
\begin{aligned}
\mathcal L_{\mathrm{OPSD}}^{(r)}
={}&-\frac{1}{Z_r}\sum_{i\in I_r}\sum_t m_{i,t}A^{\textsc{opsd}}_{i,t}\\
&\quad\cdot\log\pi_\theta^r(y_{i,t}\mid q,y_{<i},y_{i,<t}).
\end{aligned}
\end{equation}
Only ordinary-context probabilities carry gradients to $\theta_r$.
OPSD uses neither ratios nor clipping. Appendix~\ref{app:coefficient} gives
the caching \mbox{and reduction details.}

\paragraph{Joint training and linear annealing.}
Our experiments combine GRPO's clipped policy-gradient loss with OPSD,
as illustrated in \mbox{Figure~\ref{fig:method}'s upper panel:}
\begin{equation}
\begin{aligned}
\mathcal{L}^{(r)}(s)&=\mathcal{L}_{\mathrm{GRPO}}^{(r)}+
w^O(s)\mathcal{L}_{\mathrm{OPSD}}^{(r)},\\
w^O(s)&=\max(w_0-s\rho,w_{\min}).
\end{aligned}
\end{equation}
Here $s$ is the training step, while $w_0$, $\rho$, and $w_{\min}$ set the initial
weight, decay, and floor. Both terms update $\theta_r$: GRPO and LOO share
DrMAS agent-wise advantage $A_i^r$, whereas OPSD supplies fact-conditioned
token-varying credit.
Annealing reduces the OPSD contribution while keeping verified outcome
feedback active. Appendix~\ref{app:algorithm} gives implementation details and
loss aggregations. Inference uses ordinary contexts, without future
communication or the external model.

\newcommand{\ExpandedFWFourMathPOne}{91.50}
\newcommand{\ExpandedVsFrozenMathPOne}{8.00}
\newcommand{\ExpandedFWFourMathPFour}{96.60}
\newcommand{\ExpandedVsFrozenMathPFour}{4.80}
\newcommand{\ExpandedFWFourOlympiadPOne}{60.30}
\newcommand{\ExpandedVsFrozenOlympiadPOne}{13.49}
\newcommand{\ExpandedFWFourOlympiadPFour}{69.93}
\newcommand{\ExpandedVsFrozenOlympiadPFour}{10.67}
\newcommand{\ExpandedBaseAMCPOne}{60.24}
\newcommand{\ExpandedFrozenAMCPOne}{56.63}
\newcommand{\ExpandedRelaxedVsNoneHundred}{6.13}
\newcommand{\ExpandedRelaxedVsNoneHundredTen}{4.27}
\newcommand{\ExpandedRelaxedVsNoneHundredTwenty}{1.98}
\newcommand{\ExpandedRelaxedVsNoneHundredThirty}{2.10}
\newcommand{\ExpandedMatchingGapMin}{2.86}
\newcommand{\ExpandedMatchingGapMax}{8.12}

\section{Experiments}
\label{sec:experiments}

\subsection{Setup}
\label{sec:setup}

\paragraph{Models and data.}
We initialize trained agents from Qwen3-4B \citep{yang2025qwen3},
and all settings use non-thinking mode.
AFlow \citep{zhang2025aflow} and GPTSwarm \citep{zhuge2024gptswarm}
also use Qwen3-4B executors. AFlow's search optimizer is
Qwen3-235B-A22B-Instruct-2507-Pro \citep{qwen2025instruct2507}.
For policy training, we use DAPO-MATH-17k \citep{yu2025dapo}.
We test on six other datasets: AMC23, AIME24, AIME25 \citep{maaAMC},
MATH-500 \citep{lightman2023verify}, the English text-only mathematics
subset of OlympiadBench \citep{he2024olympiadbench}, and
Minerva \citep{lewkowycz2022minerva}.

\paragraph{Compared methods.}
(1) \emph{Base Model}: Qwen3-4B answers directly without multi-agent interaction.
(2) \emph{Frozen-weight MAS}: Solver--Verifier MAS uses the autonomous
interaction loop shared by DrMAS and OPSD4MAS, with model-decided termination
and no LLM parameter updates. AFlow and GPTSwarm optimize collaboration
structures or prompts while retaining fixed LLM weights.
(3) \emph{Scalar-reward RL}: DrMAS \citep{feng2026drmas} with GRPO
\citep{shao2024deepseekmath} or DAPO \citep{yu2025dapo} uses outcome rewards.
CoMAS \citep{xue2026comas} derives scalar rewards from interactions.
(4) \emph{OPSD4MAS}: FW2 and FW4 combine OPSD with GRPO, using future
windows of two and four responses.
Table~\ref{tab:main} reports pass@1 and pass@4 as percentages.
Appendix~\ref{app:planned-comparisons} gives protocols for \mbox{evaluation and selection.}

\subsection{Main Results}

\paragraph{Gains over scalar-reward RL.}
FW4 leads all six datasets in Table~\ref{tab:main}.
With equal dataset weights, mean pass@1 exceeds GRPO, DAPO, and CoMAS by
\DeltaFWFourVsGRPOSixDatasetMeanPOne{}, \DeltaFWFourVsDAPOSixDatasetMeanPOne{}, and
\DeltaFWFourVsCoMASSixDatasetMeanPOne{} percentage points. Mean pass@4 gains are
\DeltaFWFourVsGRPOSixDatasetMeanPFour{}, \DeltaFWFourVsDAPOSixDatasetMeanPFour{}, and
\DeltaFWFourVsCoMASSixDatasetMeanPFour{} percentage points, respectively.
The gains over CoMAS extend this comparison to communication-informed
rewards. Across both scalar-reward routes, these results support combining
outcome credit with communication-derived token-level credit.

\paragraph{Gains over frozen-weight MAS.}
FW4 also exceeds every reported frozen-weight result, including both the
shared Solver--Verifier loop and optimized collaboration structures.
On MATH-500, it reaches \ExpandedFWFourMathPOne{}\% pass@1 and
\ExpandedFWFourMathPFour{}\% pass@4, improving on Solver--Verifier MAS by
\ExpandedVsFrozenMathPOne{} and \ExpandedVsFrozenMathPFour{} percentage
points. On Olympiad, the respective gains are
\ExpandedVsFrozenOlympiadPOne{} and \ExpandedVsFrozenOlympiadPFour{}
percentage points. With Qwen3-4B executors throughout, these results favor
learning agent policies over relying solely on fixed-weight collaboration.

\paragraph{Benefits of a larger future window.}
FW2 is already competitive with scalar-reward RL. Across six datasets,
its mean pass@1 is close to DAPO's, while its mean pass@4 is
\DeltaFWTwoVsDAPOSixDatasetMeanPFour{} percentage points higher.
Expanding the window to four responses raises mean pass@1 and pass@4 by
\DeltaFWFourVsFWTwoSixDatasetMeanPOne{} and \DeltaFWFourVsFWTwoSixDatasetMeanPFour{}
percentage points.
The observed gains support a broader span of future
communication as a valuable source of information for fact selection.

\subsection{Ablating Future Communication Processing}

\begin{table}[!t]
\centering\small
\setlength{\tabcolsep}{2.2pt}
\begin{tabular*}{\columnwidth}{@{\extracolsep{\fill}}lcccrr@{}}
\toprule
Condition & $E$ & $A$ & $L$ & $p_1$ & $p_4$ \\
\midrule
No OPSD & -- & -- & -- & 49.63 & 66.60 \\
Raw Commun. & -- & -- & -- & 33.24 & 50.46 \\
Unfiltered Facts & $\checkmark$ & -- & -- & 42.57 & 63.97 \\
Coarse-filtered Facts & $\checkmark$ & $\checkmark$ & -- & 42.99 & 65.18 \\
Full & $\checkmark$ & $\checkmark$ & $\checkmark$ & \textbf{64.91} & \textbf{82.74} \\
\bottomrule\end{tabular*}
\caption{Training ablations: mean pass@1/pass@4 ($p_1/p_4$, \%) on AMC23, AIME24, and AIME25, with equal dataset weights. Raw Commun. abbreviates Raw future communication. $E/A/L$: extraction/Association/LOO. No OPSD and Full reuse GRPO and FW4 from Table~\ref{tab:main}.}
\label{tab:ablations}
\end{table}

Table~\ref{tab:ablations} compares five FEHF training variants.
\emph{No OPSD} uses GRPO alone. \emph{Raw future communication} supplies
unprocessed text as privileged context. \emph{Unfiltered Facts} extracts
facts without filtering. \emph{Coarse-filtered Facts} adds Association
but omits LOO, while \emph{Full} applies both filters.

Only Full exceeds No OPSD in both equal-dataset means, by
\AblFullVsBasePOne{} and \AblFullVsBasePFour{} percentage points.
Every tested incomplete variant falls below scalar-reward RL alone:
access to future communication can therefore hurt training when the
privileged context is inadequately processed.
Extraction improves mean pass@1/pass@4 over raw communication by
\AblExtractionPOne{}/\AblExtractionPFour{} percentage points, and
Association adds \AblAssociationPOne{}/\AblAssociationPFour{} percentage
points. Adding LOO produces the largest further gains,
\AblLOOPOne{}/\AblLOOPFour{} percentage points.
These results support the complete pipeline: extraction makes knowledge
individually assessable, Association prioritizes outcome-related facts,
and LOO selects facts with \mbox{favorable target-specific influence.}

\subsection{Fact Injection: Utility and Task Matching}
\label{sec:injection}
\begin{figure}[!t]
\centering
\includegraphics[width=\columnwidth]{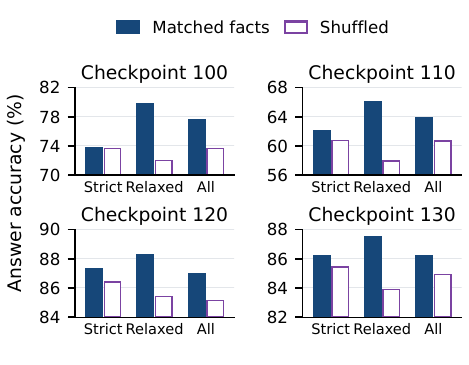}
\caption{Quantity-matched control design: Relaxed has the largest matched--shuffled gains. Nonzero y-axis ranges differ across panels, so compare numerical values across checkpoints. Protocol: Appendix~\ref{app:injection-sweep}.}
\label{fig:injection-matched}
\end{figure}

\begin{figure}[!b]
\centering
\includegraphics[width=\columnwidth]{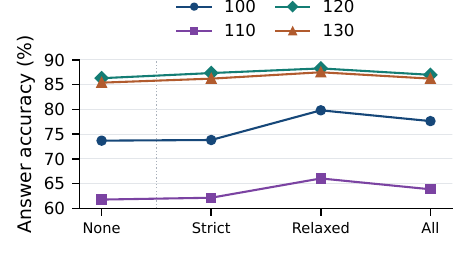}
\caption{Filtering-strength trend (60--90\% axis): Relaxed is best at all checkpoints. None is the no-injection reference, and Strict through All relax LOO. Category spacing does not encode fact quantity.}
\label{fig:injection-strength}
\end{figure}

\begin{figure*}[t]
\centering
\includegraphics[width=\textwidth]{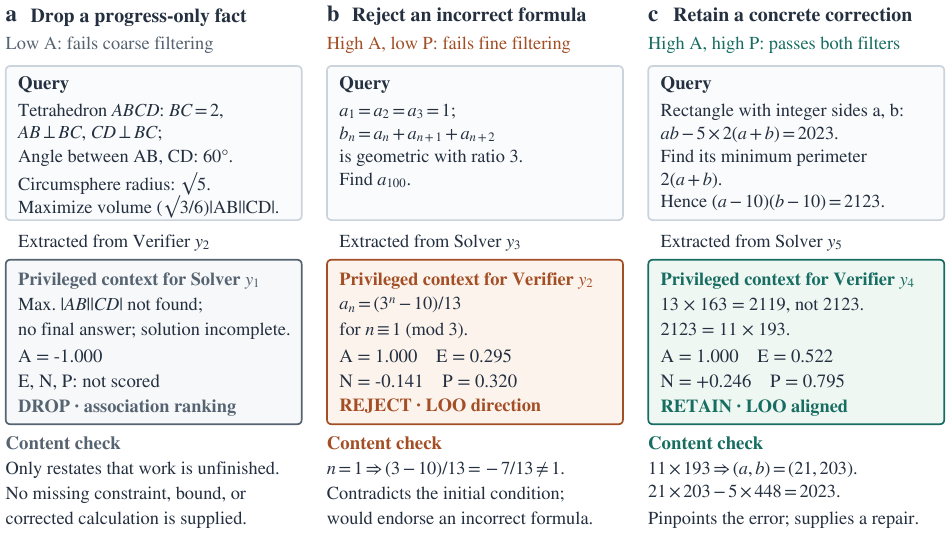}
\caption{Selected cases connect filtering decisions to fact content:
(a) no actionable repair, (b) a falsifiable error, (c) a concrete correction.
Queries and facts are abridged. Boxed facts are proposed privileged contexts
for the named responses. DROP/REJECT denotes non-retention, not injection.
A/E/N/P denote association, RMS effect, net alignment,
and directional precision. Content checks are independent of logged
selection, with sources and scores in Appendix~\ref{app:fig4-cases}.}
\label{fig:fact-cases}
\end{figure*}

We vary LOO filtering strength and inject the resulting facts as privileged
context to assess their downstream utility. At four FW4 checkpoints, we
freeze saved histories and regenerate only the final active Solver on
training questions. Facts enter historical and final-input slots.
\emph{None} supplies no facts, while \emph{Strict} and \emph{Relaxed} apply
different LOO thresholds. \emph{All} bypasses LOO on the
\mbox{post-association candidate pool.}

\paragraph{Task matching under quantity controls.}
For each filtering profile, Shuffled replaces the original facts with
facts from other questions. The paired design keeps injection slots and
visible fact counts the same, with approximate length matching, to test
whether appropriate content helps beyond adding more text.
As shown in Figure~\ref{fig:injection-matched}, Relaxed yields the largest
matched--shuffled advantage at every checkpoint,
\ExpandedMatchingGapMin{}--\ExpandedMatchingGapMax{} percentage points.
This contrast supports task-appropriate selection, rather than extra text
alone, as a \mbox{source of benefit.}

\paragraph{An intermediate filtering strength works best.}
Figure~\ref{fig:injection-strength} compares the ordered settings
None, Strict, Relaxed, and All. Relative to the no-injection baseline,
Relaxed improves accuracy by
\ExpandedRelaxedVsNoneHundred{}/\ExpandedRelaxedVsNoneHundredTen{}/\ExpandedRelaxedVsNoneHundredTwenty{}/\ExpandedRelaxedVsNoneHundredThirty{}
percentage points at checkpoints 100/110/120/130.
Accuracy rises from Strict to Relaxed and falls from Relaxed to All at
every checkpoint. Neither very restrictive selection nor removing the LOO
gate gives the best result: useful privileged context requires an
appropriate filtering strength, not simply a larger fact pool.

\subsection{Case Study: Understanding Fact Selection}
\label{sec:analysis}

We examine representative facts to explain what our filters remove and
retain. Figure~\ref{fig:fact-cases} connects each decision to the fact's
content, source response, and proposed target.

\paragraph{Low association: no actionable repair.}
Panel (a) asks for a tetrahedron's maximum volume. The fact extracted from
Verifier $y_2$ tells Solver $y_1$ only that the optimization is unfinished.
It supplies no missing constraint, bound, or corrected calculation.
With $A=-1$, this progress description is dropped before LOO scoring.

\paragraph{High association, unfavorable local direction.}
Panel (b) specifies $a_1=a_2=a_3=1$ and geometric consecutive triple sums
with ratio 3. Solver $y_3$ supplies Verifier $y_2$ with the claim
$a_n=(3^n-10)/13$ for $n\equiv1\pmod3$. At $n=1$, it gives $-7/13$,
contradicting the initial condition. Despite $A=1$ and substantial effect
$E=0.295$, negative alignment $N=-0.141$ and low directional precision
$P=0.320$ lead to rejection. Strong teacher influence alone is insufficient, so
the local screen also tests its direction relative to outcome credit.

\paragraph{High association, favorable local direction.}
Panel (c) factors $(a-10)(b-10)=2123$ to minimize a rectangle's
perimeter. The fact from Solver $y_5$ for Verifier $y_4$ corrects
$13\times163=2119$ and supplies $2123=11\times193$, yielding sides 21 and
203 and perimeter 448. The repair is retained with $A=1$, $E=0.522$,
$N=0.246$, and $P=0.795$.

\begingroup\widowpenalty=10000
These cases illustrate how FEHF turns future communication into privileged
context. In particular, LOO selects facts by their token-level influence
relative to the target's agent-wise GRPO advantage. The annealed OPSD--RLVR
objective then combines agent-wise outcome credit with fact-conditioned
token-level credit.
\par
\endgroup

\section{Conclusion}

We observe that scalar-reward MAS training compresses or ignores communication
and assigns coarse credit. To address this limitation, we propose \method{},
using Fact Extraction and Hierarchical Filtering (FEHF) to turn future
communication into privileged context. Its linearly annealed joint objective
combines OPSD token-level credit with GRPO outcome guidance. Supporting this
design, OPSD4MAS with a future window of four responses (FW4) leads all
baselines across six mathematical benchmarks. Ablations support
complete FEHF, while injection diagnostics highlight task matching and
filtering strength. Together, FEHF converts future communication into
privileged context, enabling annealed OPSD--RLVR joint training to combine
agent-wise outcome credit with fact-conditioned token-level credit.

\section*{Limitations}

Our evaluation focuses on mathematical reasoning with Qwen3-4B agents.
Extending the approach to other tasks, model scales, and interaction
structures remains future work. Fact extraction and relation labeling rely
on an external model, so the quality of privileged context can depend on
that model and on filtering thresholds, \mbox{context budgets, and annealing.}

The injection diagnostic evaluates selected facts at fixed weights and
frozen histories. Its four checkpoints belong to one training run rather
than independent training seeds, so the observed checkpoint differences
do not measure \mbox{variation across training runs.}

\section*{Ethical Considerations}

This work studies reinforcement learning on competition-mathematics data and
does not use personal or sensitive information. External model calls receive
problem-solving traces, so deployments should ensure that traces contain no
confidential data and comply with the provider's data policy. More broadly,
multi-agent self-improvement can amplify incorrect feedback shared between
agents. The proposed filters are intended to mitigate this risk, but a reduction
has not been demonstrated. Selected facts should not be interpreted as
verified truth outside the evaluated domain.

OpenAI Codex was used to assist with code development and debugging,
language revision, figure prototyping, and consistency checking.
All AI-assisted material included in this submission was reviewed by the
authors, who remain fully responsible for the content, claims, citations,
code, and reported results.

\bibliography{references,planned_references,setup_references}
\appendix
\section{Implementation and Reproducibility}
\label{app:algorithm}

\paragraph{Computing environment.}
The project used a computing environment with 24 NVIDIA A100 GPUs,
each with 80\,GB of memory. This describes the available project resources,
not the allocation of every individual run.

\paragraph{Baseline objectives and parameter indexing.}
\label{app:baseline-objectives}
DrMAS uses a clipped importance-ratio objective for each agent, with outcome
credit normalized by agent action statistics \citep{feng2026drmas}.
We follow its query-and-agent response grouping in
Equation~\ref{eq:grpo-outcome-advantage}.
CoMAS instead collects agent outputs and interaction-derived rewards into
a buffer and optimizes a REINFORCE++ clipped surrogate \citep{xue2026comas}.
For output $o$ and input $p$, its pre-normalization advantage includes
a cumulative KL term,
\begin{equation}
A_t=r(o)-\beta\sum_{u=t}^{|o|}
\log\frac{\pi_{\theta_k}(o_u\mid p,o_{<u})}
{\pi_{\mathrm{ref}}(o_u\mid p,o_{<u})}.
\end{equation}
After batch normalization, the objective sums
$\min(\rho_t\widehat A_t,\operatorname{clip}(\rho_t,1-\epsilon,1+\epsilon)
\widehat A_t)$ over tokens, with current/old-policy ratio $\rho_t$.
The reported zero-KL-weight configuration has a token-constant reward
component. CoMAS agents can take different interaction roles, whereas
$\theta_r$ in our formulation indexes a fixed role policy.

\paragraph{Cached scores and loss reduction.}
\label{app:coefficient}
Teacher and ordinary-context token scores are cached before actor updates,
and their difference is computed without gradient tracking. The OPSD
coefficient is detached, so gradients flow only through ordinary-context
token log probabilities in the actor loss. Extraction and filtering do not
receive gradients.
The default OPSD loss averages over active response tokens in each
role-homogeneous worker microbatch before gradient accumulation. Prompt
and inactive tokens are excluded. GRPO instead averages each response's
masked token mean over nonempty responses of that role and uses the clipped
ordinary-policy likelihood ratio. OPSD uses neither ratios nor clipping.
These distinct reductions are retained in the joint objective.

\paragraph{Two outcome advantages.}
Association computes $A(\tau_j)$ from one episode reward per trajectory
within query $q$. Its statistics use trajectories identified by the query's
relation maps, with population standard deviation and $\epsilon=10^{-6}$.
Zero-variance groups, including singletons, receive zero.
These values weight fact--trajectory relations rather than the actor loss.

GRPO computes the DrMAS agent-wise advantage from response rewards
$s_{j,i}=\sum_t r_{j,i,t}$, grouping responses by query and agent ID.
Repeated turns count separately. For at least two responses, normalization
uses the sample standard deviation with denominator $n-1$.
A singleton uses mean zero and standard deviation one. Equal-reward groups
of at least two responses receive zero.
Association uses episode rewards, while GRPO uses the trainer's response
reward tensor, including any configured reward processing.
LOO reads the same cached \texttt{advantages\_grpo} as the actor loss rather
than recomputing an Association advantage.

\paragraph{Candidate selection and ranking.}
A source fact can reach earlier active non-judger responses in the same
trajectory within $W$ positions. Validated claims from entirely unsuccessful
groups are excluded, and duplicate normalized lesson text is collapsed.
Candidate ranking uses descending Association, relation coverage, and type
priority (correction, warning, validated claim, strategy), followed by
shorter future distance and ascending fact ID. Duplicate ties retain the
larger ID. Facts passing LOO are ranked by descending net alignment,
precision, and RMS effect, then shorter distance and ascending ID before
top-$K$ truncation. All LOO token averages use active response masks.
The final selected set receives a fresh teacher-forced scoring pass for OPSD.

\paragraph{Teacher-context construction.}
\label{app:serialized-contrast}
The serializer appends [Privileged Future Lessons] and numbered facts to
the last user message, then applies the chat template with a generation
marker. To meet the context budget, it first removes lessons from the end
of their supplied order. If necessary, it removes older unprotected history
messages while retaining the first message, first user message, and latest
message, then removes further lessons.
Full-candidate and leave-one-out prompts are serialized separately.
Their difference isolates a single fact when the retained history and all
other facts match. Under truncation, it reflects the resulting serialized
contexts. For a single candidate, an unavailable removal score is replaced
by the cached ordinary score. A missing full score or a missing removal
score for multiple candidates causes rejection. An unavailable selected
teacher score leaves the OPSD coefficient zero.

\section{Supplementary Results and Evaluation Protocols}
\label{app:planned-controls}
\subsection{Benchmark and Baseline Protocols}
\label{app:planned-comparisons}
Agents initialize from Qwen3-4B and train on DAPO-MATH-17k.
All settings use non-thinking mode. FW2 and FW4 denote future windows
of two and four responses. Trained checkpoints are selected by validation
and then evaluated on the test sets.

\paragraph{Dataset coverage and aggregation.}
Table~\ref{tab:main} reports all nine methods on AMC23, AIME24, AIME25,
MATH-500, Olympiad, and Minerva. Main-result means weight the six datasets
equally, whereas ablation means use the three named competition datasets.
MATH-500 is the held-out subset used by \citet{lightman2023verify}, not the
complete MATH test set \citep{hendrycks2021math}. Olympiad denotes
OlympiadBench's English text-only mathematics subset
\citep{he2024olympiadbench}. Minerva follows the mathematical evaluation
setting of \citet{lewkowycz2022minerva}.
The reported Base evaluation covers 83, 30, 30, 500, 675, and 272 questions,
respectively. Accuracy is reported in percent, and differences are
percentage points. A pass@4 sample is a complete trajectory producing a
final answer, not an individual agent response within a shared interaction.

\paragraph{Fixed-weight comparisons.}
Base Model answers directly. Solver--Verifier MAS retains the initial
weights in the model-terminated interaction loop shared by DrMAS and
OPSD4MAS. AFlow and GPTSwarm use Qwen3-4B executors and optimize
collaboration structures or prompts without LLM parameter updates.
AFlow's search optimizer uses the service identifier
Qwen3-235B-A22B-Instruct-2507-Pro, with the underlying published model
documented by the Qwen team \citep{qwen2025instruct2507}.
The service suffix is retained for reproducibility.
All baseline scores are evaluations in this study rather than numbers
copied from the original method papers.

\subsection{Dataset-Level Ablation Results}
\label{sec:mechanism-tests}
Table~\ref{tab:ablation-details} gives the dataset-level results underlying
Table~\ref{tab:ablations}. Each metric averages AMC23, AIME24, and AIME25
equally, using full supplied precision before rounding.
No OPSD and Full reuse the DrMAS-GRPO and OPSD4MAS (FW4) results from
Table~\ref{tab:main}, respectively. The intermediate settings compare raw
future communication, extraction alone, and extraction with Association.
\begin{table*}[t]
\centering\small
\begin{tabular}{lrrrrrr}
\toprule
Condition & \multicolumn{2}{c}{AMC23} & \multicolumn{2}{c}{AIME24} & \multicolumn{2}{c}{AIME25} \\
 & $p_1$ & $p_4$ & $p_1$ & $p_4$ & $p_1$ & $p_4$ \\
\midrule
No OPSD & 68.07 & 83.13 & 48.33 & 63.33 & 32.50 & 53.33 \\
Raw future communication & 57.22 & 74.70 & 25.00 & 46.67 & 17.50 & 30.00 \\
Unfiltered Facts & 63.55 & 81.92 & 40.83 & 66.67 & 23.33 & 43.33 \\
Coarse-filtered Facts & 65.63 & 85.54 & 34.17 & 63.33 & 29.17 & 46.67 \\
Full & 82.23 & 91.57 & 62.50 & 83.33 & 50.00 & 73.33 \\
\bottomrule\end{tabular}
\caption{Per-dataset ablation results (\%). Displayed values are rounded, whereas Table~\ref{tab:ablations} averages the original supplied precision, not these rounded cells.}
\label{tab:ablation-details}
\end{table*}

\subsection{Four-Checkpoint Injection Diagnostic}
\label{app:injection}
\label{app:injection-sweep}

\paragraph{Evaluation setting.}
Figures~\ref{fig:injection-matched} and \ref{fig:injection-strength} use FW4
checkpoints 100, 110, 120, and 130 with their saved trajectories and facts.
The evaluator freezes the MAS history, inserts facts at targeted historical
positions and the final Solver input, and regenerates only the final active
Solver on training questions. Model weights remain fixed during this diagnostic.

\paragraph{Filtering profiles.}
Strict requires RMS effect $E\geq0.10$, directional precision $P\geq0.65$,
and net alignment $N>0.001$. Relaxed uses $E\geq0.005$, $P\geq0.50$,
and $N>-0.001$, with configuration name \texttt{loose}.
All uses the post-association candidate pool without a LOO gate, including
candidates with missing LOO scores. It does not restore Association-rejected
facts. None is the no-injection reference.

\paragraph{Matched controls and statistical unit.}
For each profile, Shuffled uses facts from other questions at corresponding
injection slots, matching visible fact counts with approximate type and
length matching. It does not sample random facts from the current question.
Queries are the independent units for statistical inference, with trajectories
and repeated answers nested within them. The four checkpoints are from one
training run. Comparisons are descriptive, without significance tests.

\paragraph{Figure scales.}
Figure~\ref{fig:injection-matched} uses checkpoint-specific nonzero ranges
of 70--82\%, 56--68\%, 84--90\%, and 82--88\%.
Numerical values, rather than bar heights across panels, should be compared.
Figure~\ref{fig:injection-strength} uses 60--90\%.
Its category spacing does not represent equal fact-count increments.

\begin{table*}[t]
\centering\small
\begin{tabular}{llllll}
\toprule
Panel / ID & Step & Fact ID & Fact / traj. line & Source / target & Decision\\
\midrule
(a) / B3 & 130 & \texttt{3a722f49767d87d137954ec5} & 500 / 230 & 2 / 1 & Association rejected\\
(b) / C1 & 130 & \texttt{4c0f1ebe82fad73ec9e52a59} & 652 / 248 & 3 / 2 & LOO rejected\\
(c) / A1 & 130 & \texttt{3fa5b26b4d135ef12d49acb0} & 558 / 15 & 5 / 4 & LOO aligned\\
\bottomrule
\end{tabular}
\caption{Sources of the cases in Figure~\ref{fig:fact-cases}. Line numbers
index the fact-card and trajectory exports from one. Source/target gives
the response positions within each trajectory.}
\label{tab:fig4-provenance}
\end{table*}

\section{Extractor and Relationer Prompts}
\label{app:fact-prompts}
These literal templates are exported from the implementation.
Braced uppercase slots denote runtime substitutions, not example data.
The accompanying source bundle records the builder path and SHA256 hash.
\paragraph{Extractor input.}
RECORDS\_JSON is an array of records containing \texttt{record\_id},
\texttt{problem}, \texttt{evaluated\_response}, and \texttt{comment}.
MAX\_FACTS and EXTRACTION\_VERSION come from configuration.
\begin{quote}\raggedright
You are a mathematical Fact Card extractor. Analyze every record independently and output only the most decision-relevant mathematical lessons. The comment is fallible evidence,\allowbreak{} not an authority:\allowbreak{} verify the evaluated\_response against the problem yourself,\allowbreak{} even when the comment approves it or fails to mention an error. When comment is empty,\allowbreak{} analyze evaluated\_response directly. Identify the decisive mathematical pivot,\allowbreak{} in both directions:\allowbreak{} the earliest invalid step that can make the final answer wrong,\allowbreak{} and the key successful step or result that materially makes the correct answer possible. A decisive success factor may be the right modeling reduction,\allowbreak{} theorem,\allowbreak{} invariant,\allowbreak{} formula,\allowbreak{} substitution,\allowbreak{} construction,\allowbreak{} bound,\allowbreak{} or feasibility constraint; prefer the smallest concrete step without which the conclusion would not follow,\allowbreak{} rather than routine algebra or a summary of the whole solution. Check formulas,\allowbreak{} theorem hypotheses,\allowbreak{} algebraic substitutions,\allowbreak{} signs,\allowbreak{} coefficients,\allowbreak{} arithmetic,\allowbreak{} domains,\allowbreak{} quantifiers,\allowbreak{} integrality or positivity constraints,\allowbreak{} and whether the conclusion is necessary and sufficient. Preserve exact equations and explicitly contrast an incorrect formula or substitution with its correct replacement. Do not merely restate the final answer. Order facts by priority:\allowbreak{} (1) decisive correction of the earliest root-cause error and decisive validated success factor or key step; (2) supporting verified formula,\allowbreak{} theorem,\allowbreak{} or derivation; (3) specific warning or validation check; (4) general strategy only when no more concrete mathematical lesson is available. Use fact\_type=correction for a specific mathematical error and its fix; validated\_claim only for a claim independently verified as correct,\allowbreak{} especially a decisive success-producing step; warning for a concrete failure mode; strategy for a specific reusable method. Each lesson must be atomic,\allowbreak{} actionable,\allowbreak{} self-contained,\allowbreak{} concise,\allowbreak{} and usable without the original response. Reject bare correctness verdicts,\allowbreak{} vague advice such as 'be rigorous',\allowbreak{} duplicated lessons,\allowbreak{} and unsupported claims. If no reliable mathematical lesson exists,\allowbreak{} return an empty facts list. Return at most \{\allowbreak{}MAX\_FACTS\}\allowbreak{} facts per record in priority order. Return strict JSON only with schema \{\allowbreak{}"records":\allowbreak{}[\allowbreak{}\{\allowbreak{}"record\_id":\allowbreak{}"...",\allowbreak{}"facts":\allowbreak{}[\allowbreak{}\{\allowbreak{}"lesson\_text":\allowbreak{}"...",\allowbreak{}"fact\_type":\allowbreak{}"correction|\allowbreak{}warning|\allowbreak{}validated\_claim|\allowbreak{}strategy"\}\allowbreak{}]\allowbreak{}\}\allowbreak{}]\allowbreak{}\}\allowbreak{}. Version:\allowbreak{} \{\allowbreak{}EXTRACTION\_VERSION\}\allowbreak{}\par 
Records:\allowbreak{}\par 
\{\allowbreak{}RECORDS\_JSON\}\allowbreak{}
\end{quote}
\paragraph{Relationer input.}
FACTS\_JSON contains \texttt{fact\_id} and \texttt{lesson\_text}.
TRAJECTORIES\_JSON contains trajectory identifiers and ordered responses.
RELATION\_VERSION \mbox{comes from configuration.}
\begin{quote}\raggedright
For each lesson,\allowbreak{} classify each trajectory's observable behavior. followed means clear consistent behavior without a clear opposite; violated means clear opposite behavior not clearly corrected; unrelated means absent,\allowbreak{} insufficient,\allowbreak{} or mixed behavior. Do not make causal claims. Return strict JSON only with schema \{\allowbreak{}"facts":\allowbreak{}[\allowbreak{}\{\allowbreak{}"fact\_id":\allowbreak{}"...",\allowbreak{}"relations":\allowbreak{}\{\allowbreak{}"trajectory\_uid":\allowbreak{}"followed|\allowbreak{}violated|\allowbreak{}unrelated"\}\allowbreak{}\}\allowbreak{}]\allowbreak{}\}\allowbreak{}. Version:\allowbreak{} \{\allowbreak{}RELATION\_VERSION\}\allowbreak{}\par 
Facts:\allowbreak{}\par 
\{\allowbreak{}FACTS\_JSON\}\allowbreak{}\par 
Trajectories:\allowbreak{}\par 
\{\allowbreak{}TRAJECTORIES\_JSON\}\allowbreak{}
\end{quote}

\section{Fact-Selection Case Sources}
\label{app:fig4-cases}

The three cases in Figure~\ref{fig:fact-cases} come from the FW4
checkpoint-130 export. Table~\ref{tab:fig4-provenance} identifies the source
records and decisions. Question text comes from saved student prompts.
The accompanying records preserve full fact text, response identifiers,
target scores, and input hashes.

\paragraph{Content checks and recorded scores.}
In (a), the fact describes unfinished optimization without giving a missing
constraint or calculation. Its Association is approximately $-1$, and the
target is rejected before LOO scoring.
In (b), the proposed recurrence yields $a_1=-7/13$ rather than the required
$a_1=1$. Its recorded scores are
$(A,E,N,P)\approx(1,0.294626,-0.140895,0.320246)$.
In (c), $11\times193=2123$ gives sides 21 and 203 with perimeter 448,
and direct substitution verifies $21\times203-5\times448=2023$.
The recorded scores for its fourth-response target are
$(A,E,N,P)\approx(1,0.521616,0.245840,0.794624)$.
These mathematical checks accompany the logged filtering decisions,
not additional regenerated answers.

\end{document}